# Feature-Based TAG in place of multi-component adjunction: Computational Implications

B.A. Hockey
Dept. of Linguistics
University of Pennsylvania
Philadelphia PA 19104.
email: beth@linc.cis.upenn.edu

B. Srinivas
Dept. of Computer and Info. Science
University of Pennsylvania
Philadelphia PA 19104.
email: srini@linc.cis.upenn.edu

**Abstract**

Using feature-based Tree Adjoining Grammar (TAG), this paper presents linguistically motivated analyses of constructions claimed to require multi-component adjunction. These feature-based TAG analyses permit parsing of these constructions using an existing unification-based Earley-style TAG parser, thus obviating the need for a multi-component TAG parser without sacrificing linguistic coverage for English.

## 1 Introduction

It has been argued that the analysis of certain linguistic constructions requires an extension of the basic tree-adjoining grammar (TAG) formalism to include multi-component adjunction [4, 3]. The restricted version of multi-component adjunction suggested for these constructions does not change the weak or strong generative capacity of the formalism. This paper demonstrates how these constructions can be handled using feature-based TAGs, thereby eliminating the need to construct a parser for TAGs with multi-component adjunction. This would make it possible to parse such constructions with the current implementation of the feature-based TAG parser [7].

Our analysis first develops the alternative suggested by Kroch and Joshi (1987)[4] for handling extraposition with features and then extends the approach to the other cases in English that appear to require multi-component adjunction, such as extraction from PP adjuncts and extraction from indirect questions. The feature based TAG analyses for these cases are as linguistically well-motivated as analyses that require multi-component adjunction (e.g. [4]).

## 2 Tree Adjoining Grammar (TAG) formalism

The analysis in this paper is based on two extensions to the TAG formalism developed in Joshi, Levy, Takahashi (1975) [2]: feature structures [8] and multi-component adjunction [2, 3, 4]. The reader is referred to the references cited in this section for more detailed discussion of the formalism than will be provided in this paper.

The primitive elements of the TAG formalism, ELEMENTARY TREES, are of two types: INITIAL TREES and AUXILIARY TREES. In a TAG grammar for natural language, INITIAL TREES are phrase structure trees of simple sentences containing no recursion, while recursive structures are represented by AUXILIARY TREES. Examples of initial and auxiliary trees are shown in Figure 1. Nodes on the frontier of initial trees are marked as substitution sites by a ($\downarrow$), while exactly one node on the frontier of an auxiliary tree whose label matches the label of the root of the tree, is marked as a foot node by a ($*$). The other nodes on the frontier of an auxiliary tree are marked as substitution sites. The elementary trees define the domain of locality

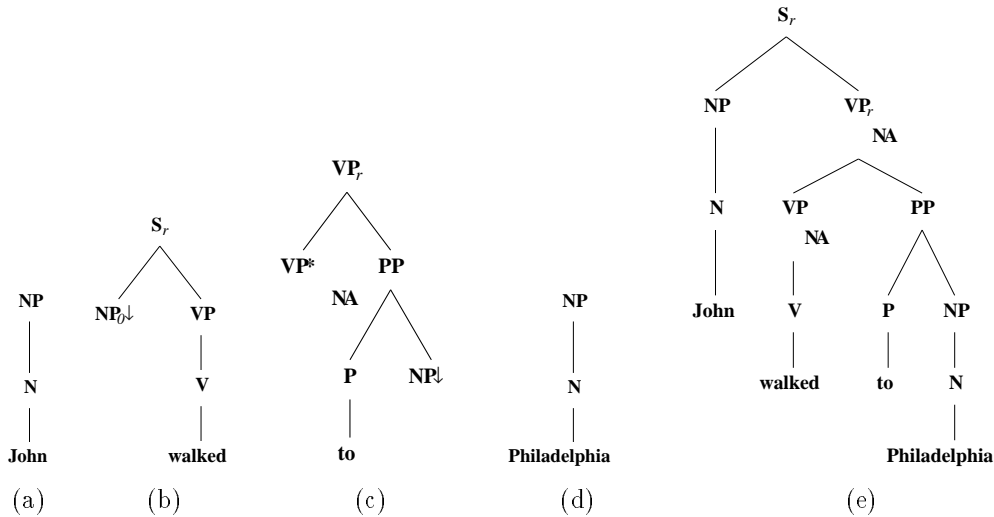

Figure 1: Elementary Trees for *John walked to Philadelphia*

over which constraints are specified.

Elementary trees are combined by operations of SUBSTITUTION and ADJUNCTION. Substitution inserts elementary trees into substitution nodes that appear on the frontier of other elementary trees. Adjunction grafts auxiliary trees into elementary trees at the node whose label is the same as root label of the auxiliary tree. As an example, the component trees shown in Figure 1 can be combined to form the sentence *John walked to Philadelphia* as follows:

1. Figure 1(a) substitutes at the $NP_0$ node of Figure 1(b).

2. Figure 1(d) substitutes at the NP node of Figure 1(c).

3. The result of step (2) above adjoins to the VP node of the result of step (1). The final result is shown in Figure 1(e).

The trees that can be adjoined at a node can be constrained by specifying one of the following adjoining constraints at that node.

- Selective Adjoining (SA): Only a specified subset of all the auxiliary trees is adjoinable at the node.

- Obligatory Adjoining (OA): At least one of all the auxiliary trees must be adjoined at the node.

- Null Adjoining (NA): No auxiliary tree is adjoinable at the node (node marked by NA).

Feature structures can be added to the basic TAG formalism [8, 9] by associating a top and a bottom feature structure with each node. While the top feature structure at a node expresses the constraints specified by the structure above the node, the bottom feature structure expresses the constraints specified by the subtree associated with the node.

When adjunction is performed at a node, the node "splits" and the features on the resultant tree are formed as shown in the schematic Figure 2 below.

Creating a feature clash between top and bottom features of a node is equivalent to putting an OA constraint at that node. At the end of a derivation the top and bottom features of all nodes must unify. Potential feature unification failures due to a incompatibility between the top and bottom feature values of a node can be averted by performing adjunction at that node. Adjoining an auxiliary tree whose root has features compatible with the top of the node and whose foot has features compatible with the bottom of the node will separate the conflicting top and bottom features thereby preventing unification failure at the node.

Multi-Component adjunction extends the basic formalism by having sets[1] of trees rather than single trees. There are several ways of defining the composition of tree-sets [10]. One of them is tree-local composition. Tree-local composition requires all members of a tree-set to adjoin or substitute on to the same elementary tree. The addition of tree-local multi-component adjunction does not increase the generative capacity of the formalism [4]. This paper demonstrates the use of feature-based TAGs to simulate tree-local multi-component adjunction using linguistically motivated features. With the feature-based analysis, the constructions which seemed to require multi-component adjunction can be parsed with the current implementation of the feature-based TAG parser.

---

[1] We will continue to use the term 'set' in this paper for

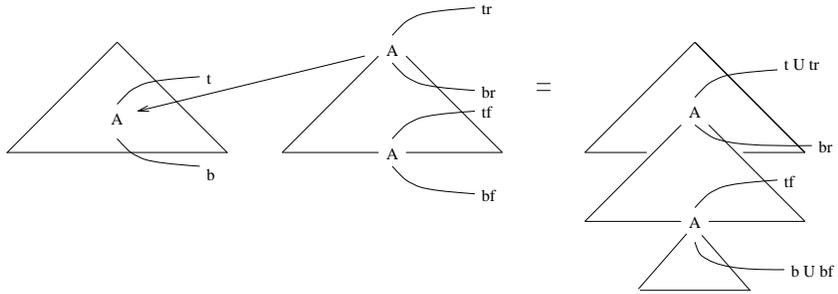

Figure 2: Schemata for feature formation upon adjunction
[t=top b=bottom r=root f=foot U=unification]

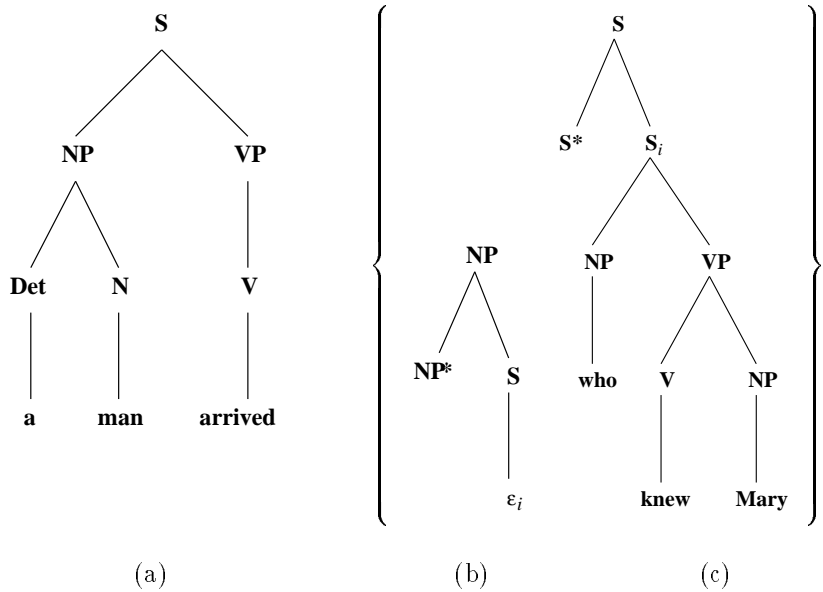

Figure 3: Component trees for *A man $\epsilon_i$ arrived [who knew Mary]$_i$*
in Kroch and Joshi's analysis

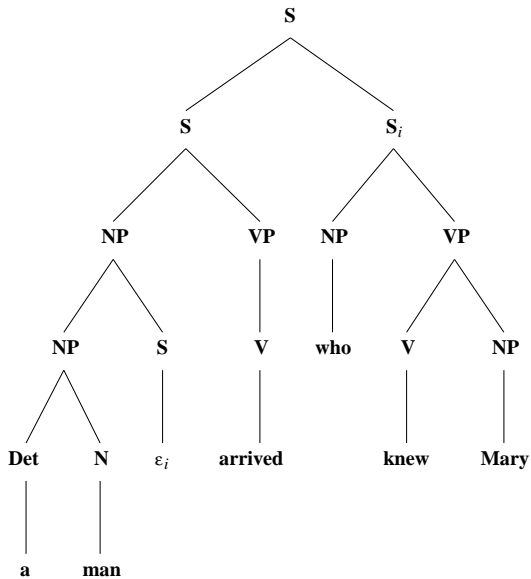

Figure 3(d): Derived tree for *A man $\epsilon_i$ arrived [who knew Mary]$_i$*
in Kroch and Joshi's analysis

## 3  Extraposition

Joshi and Kroch (1987) [4] propose a multi-component adjunction analysis for extraposition, which is illustrated by (1) below.

(1)  A man $\epsilon_i$ arrived [who knew Mary]$_i$

Their analysis requires a two-member tree-set containing the extraposed constituent, and the empty category to which the extraposed constituent is coindexed. For example, the auxiliary tree-set would be as in Figure 3(b) and Figure 3(c). The members of the tree-set would adjoin to the initial tree in Figure 3(a) to form the derived tree shown in Figure 3(d).

Our analysis for this type of construction simulates multi-component adjunction using linguistically motivated features. The feature-based TAG analysis requires three elementary trees, shown in Figure 4.

The initial tree is the one needed for the simple sentence *A man arrived.* This tree is augmented with features **displ_const** and **displ_const_index** at the root node $S_r$ and subject node $NP_0$. The value of the **displ_const** feature at a node indicates if the node dominates a trace of a displaced constituent. The **displ_const_index** feature identifies the trace with the extraposed element.

In the initial tree, the **displ_const** feature has no value, and is only coindexed between the top of $NP_0$ and the bottom of $S_r$ nodes. The substitution of *a man* at $NP_0$ instantiates <**displ_const** = -> on the bottom of $NP_0$, correctly representing the fact that the NP, *a man*, does not contain a trace of a displaced constituent. If the derivation stops at this stage the unification of the top and bottom features of each node will result in both $NP_0$ and $S_r$

---

convenience although the term 'sequence' is more accurate. In theory, the grouping in question could consist of two instances of the same tree, where both instances were required and could not viewed as a one member set. In practice this problematic instance does not seem to occur.

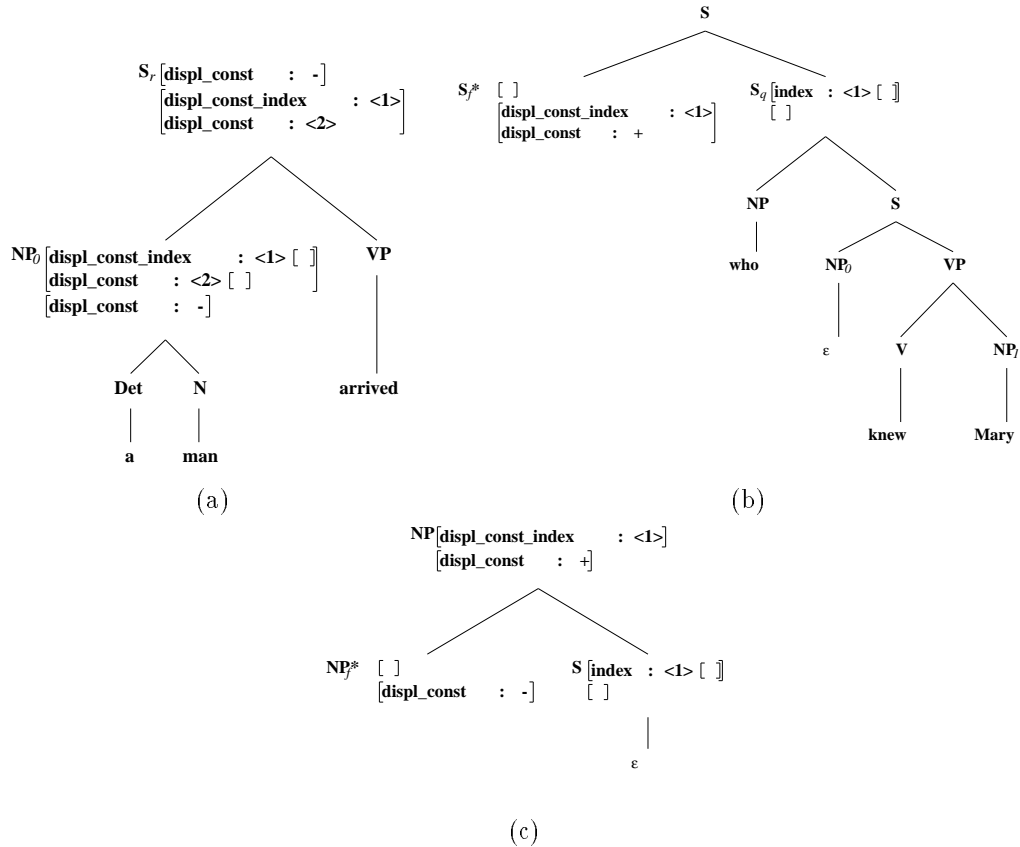

Figure 4: Component trees for *A man $\epsilon_i$ arrived [who knew Mary]$_i$* in our analysis

having <**displ_const = -**>. However, in the case of extraposition, either the extraposed clause represented by the tree in Figure 4(b) or the trace, represented by the tree in Figure 4(c) will adjoin to the tree in Figure 4(a). Either of these two adjunctions alone will introduce <**displ_const = +**> into the initial tree, resulting in a feature unification failure with <**displ_const = -**> that is present in the initial tree. The site of the unification failure is decided by the order of adjunction of these trees to the initial tree. If Figure 4(b) adjoins to $S_r$ node then the unification failure occurs between the top and the bottom feature structures at the $NP_0$ node. If Figure 4(c) adjoins to $NP_0$ node then the unification failure occurs between the top and the bottom feature structures at the $S_r$. This correctly represents the linguistic fact that with only the extraposed clause or only the trace, the resultant tree is incomplete and the derivation should not be accepted. If both the components adjoin, the resultant tree contains no feature unification conflicts and the derivation may be accepted.

Feature unification failures at a node can be remedied by adjoining an auxiliary tree whose root and foot nodes have feature structures compatible with those at the site of feature unification failure. In the case where the tree with the extraposed clause adjoins on to the $S_r$ node of the initial tree, the feature unification failure at the $NP_0$ node obligatorily forces the adjunction of the tree with the needed extraction site, shown in Figure 4(c), at the $NP_0$ node. On adjoining, the conflicting feature values for **displ_const** feature are separated and are no longer required to unify, thus resolving the unification clash. The **displ_const_index** feature passes the index between the empty category and the extraposed clause through the initial tree. Note that even though the indexing eventually is between elements from different auxiliary trees, dependencies only need to be stated within elementary trees so locality is not violated. It must also be noted that the **displ_const** feature serves to ensure that both the trees (the tree with the empty category and the tree with the extraposed clause) are adjoined collectively into the initial tree.

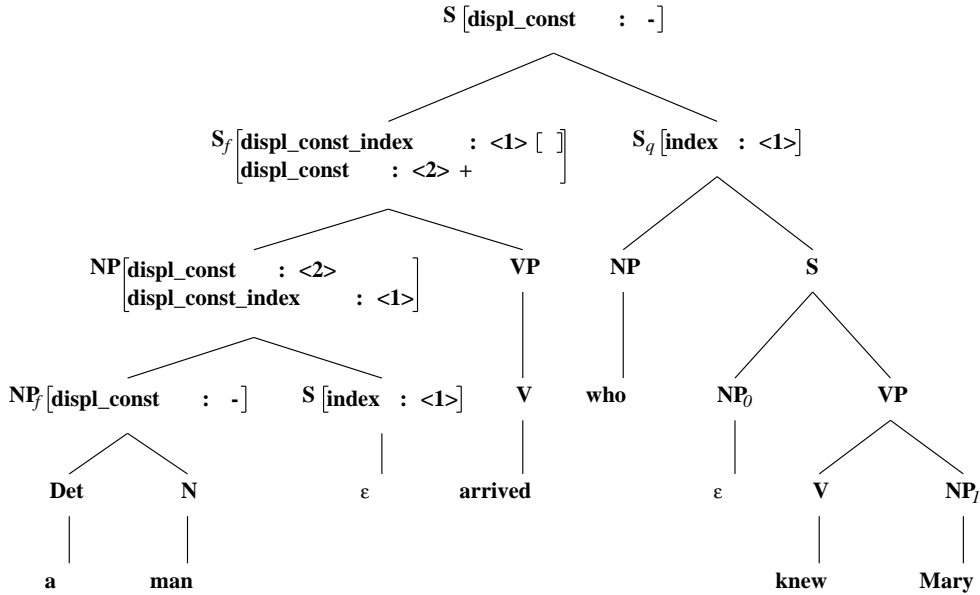

Figure 4(d): Derived tree for *A man $\epsilon_i$ arrived [who knew Mary]$_i$* in our analysis

The **displ_const** feature differs from the **ADJUNCT** feature of HPSG [6, 5] in the following way. The **displ_const** feature represents the presence of a displaced element in a given derivation and requires the introduction of the constituent to which the displaced element is related, whereas HPSG's **ADJUNCT** feature represents a list of potential adjuncts.

Extraposition of a relative clause on an object as in (2) can be handled in the same way as extraposition of relative clauses on subjects. For relative clauses on objects, the auxiliary tree containing the extraposed clause adjoins to the object NP instead of the subject NP.

(2) John gave everyone $\epsilon_i$ a hard time [who knew Mary]$_i$

In contrast to adjuncts, which must be introduced by adjunction, place-holders for arguments are present in the initial trees as substitution nodes. Therefore extraposition out of argument position, such as shown in (3), requires that the tree with the trace be substituted rather than adjoined as in the cases previously discussed. But for this difference, the analysis is similar to the analysis for adjunct extraposition.

(3) I told John $\epsilon_i$ yesterday [that I wanted pizza]$_i$. (= Kroch and Joshi (1987) [4] (50 a)).

## 4 Extraction from PP adjuncts

Extraction from PP adjuncts has been largely ignored in the literature because it has been thought to be categorically ungrammatical. However, there are examples such as (4) that are perfectly grammatical.

(4) Which gate did you leave from?

Accounting for the variation in grammaticality of extraction from adjunct PP's is beyond the scope of this paper. Rather than discuss the considerable linguistic issues raised by these constructions, we assume that such PP adjunct extractions should be included in the grammar, and propose a TAG analysis.

PP adjunct extractions are very similar to the extraposition discussed earlier. In both cases, items that are related to each other cannot be in the same initial or auxiliary tree. For extraposition, the related items were the empty category and the extraposed clause; for adjunct PP extractions the related items are the complement of the 'stranded' preposition and the extracted NP. The trees in Figure 5 show the necessary component trees for deriving example (5) below.

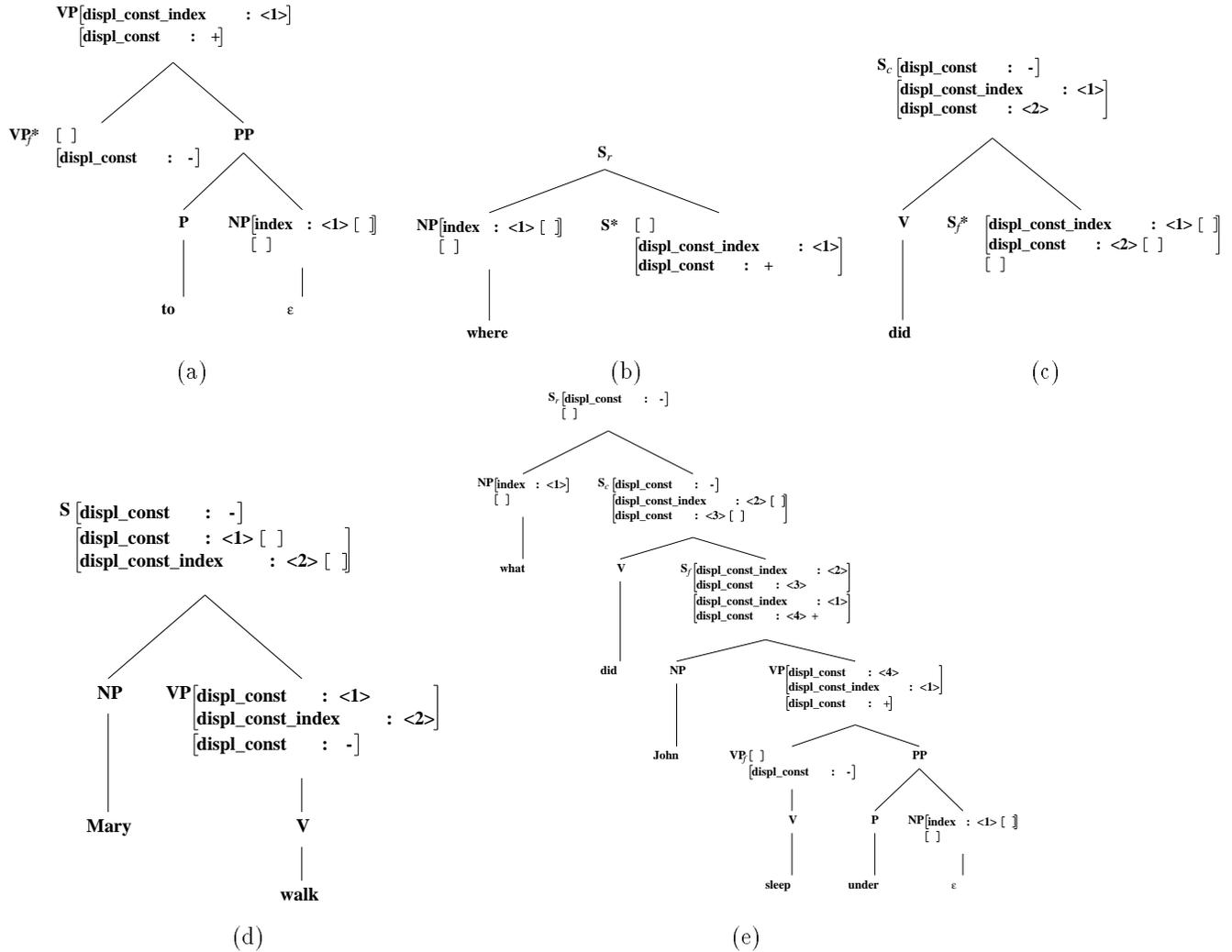

Figure 5: Component trees and the derived tree for *Where did Mary walk to ?*

(5) Where$_i$ did Mary walk to $\epsilon_i$?

The adjunction of Figure 5(a), containing the stranded preposition *to* at the VP node in Figure 5(b) introduces **displ_const** = + at the VP node and creates a feature unification failure at the S node. This forces the adjunction of Figure 5(b), containing *where* in sentence initial position, at the S node. The analysis for adjunct PP extraction is the same as that for extraposition in using the **displ_const** feature to force adjunction of the second auxiliary tree and the **displ_const_index** feature to accomplish the required coindexation.

## 5 Extraction from indirect questions

Kroch (1987) [3] argues that extractions from indirect questions, as in (6), require multi-component adjunction.

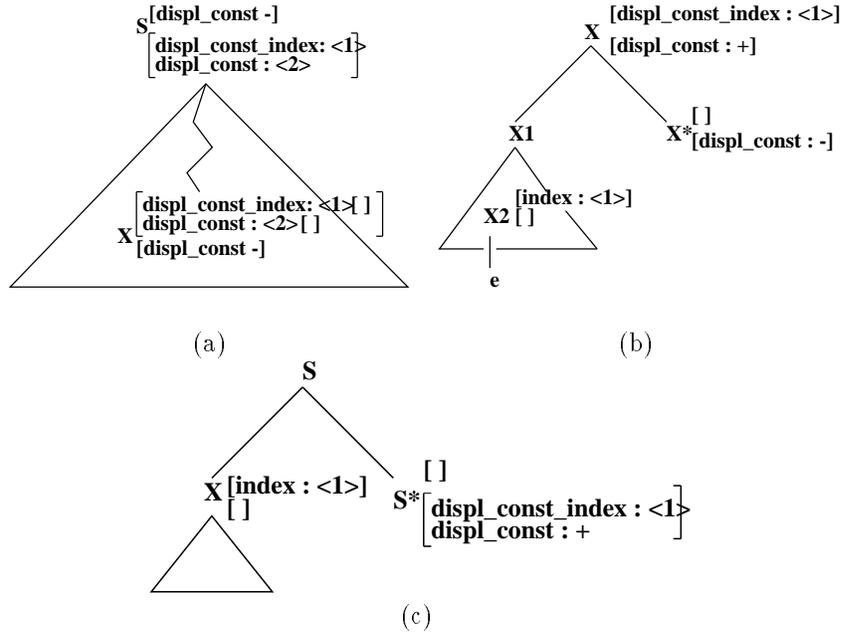

Figure 6: Schemata of trees displaying feature passing constraints

(6) (I knew) [which book]$_i$ the students would forget who$_j$ $\epsilon_j$ wrote $\epsilon_i$

Since *which book* and the empty category to which it needs to be indexed begin in separate elementary trees, Kroch uses multi-component adjunction in this case simply as a means for achieving the proper indexing. In our analysis of extraposition, we have already presented a mechanism for producing indexing between constituents that originate in different trees using the **displ_const_index** feature. Clearly a similar use of **displ_const_index** feature will work in the case of extraction from indirect questions to achieve the desired result. We omit the details for the sake of brevity.

## 6 Constraints on tree structure and feature passing

The mechanism required for simulating multi-component adjunction using the **displ_const** and **displ_const_index** features is very constrained and consistent across the phenomena we have examined. The structure of the component trees and the direction and extent of feature passing characterize the constraints. There are three types of trees involved: initial trees shown in Figure 4(a) and auxiliary trees with and without $\epsilon$ leaves shown in Figure 4(b) and Figure 4(c) respectively. Structures of initial trees are not constrained since all subcategorization possibilities must be allowed. However feature passing in initial trees is quite constrained. As can be seen in Figure 4(a), the smallest phrasal constituents of the tree have **displ_const** = - as a bottom feature and the root node has **displ_const** = - as a top feature. The **displ_const** and **displ_const_index** features are coindexed between the bottom of the root and the top of the smallest phrasal constituents. The adjunction of either type of auxiliary tree instantiates values for the coindexed features that results in obligatory adjunction of the second auxiliary tree. Auxiliary trees with the $\epsilon$ leaf consist of a root with two daughters: a foot node and a node which dominates $\epsilon$. The **displ_const_index** feature in the top feature structure of the root node is coindexed with the **index** feature of the daughter node dominating $\epsilon$. The footnode of these trees has **displ_const** = + and the bottom of the root has **displ_const** = -. The auxiliary trees that do not contain an $\epsilon$ leaf have the **displ_const_index** value coindexed between the two daughters of the root node as shown in Figure 4(c). The footnode of these trees is always S and has the feature value

displ_const = +. These tree configurations and features insure that the $\epsilon$ is always c-commanded by the constituent in the non-$\epsilon$ tree with which it is indexed.

# 7 Conclusion

English constructions that have been argued to require multi-component adjunction can be handled in a linguistically well-motivated manner by feature-based TAG analysis. In the cases of extraposition and extraction from PP adjuncts, the feature-based TAG analysis essentially simulates tree-local multi-component adjunction. For extraction from indirect questions the problem of indexing can be handled just as well by a feature-based analysis as by multi-component adjunction. We have also used the technique discussed in this paper for handling subject-auxiliary inversion. The details of the implementation are discussed elsewhere [1]. Also not discussed in this paper is the feature-based TAG analysis for extraction from recursively embedded NPs [1] which is superior to the analysis using multi-component adjunction proposed by Kroch [3]. These constructions, with the feature-based TAG analysis, can be parsed using the currently implemented unification-based TAG parser. This demonstrates that the implementational advantages of feature-based TAG can be enjoyed without any sacrifice in linguistic coverage.

# 8 Acknowledgements

The authors would like to thank Aravind K Joshi, Dania Egedi, Christy Doran, Owen Rambow, Tilman Becker for their valuable comments.